\pdfoutput=1

\documentclass[11pt]{article}

\usepackage[]{ACL2023}

\usepackage{times}
\usepackage{latexsym}
\usepackage[T1]{fontenc}
\usepackage[utf8]{inputenc}
\usepackage{microtype}

\usepackage{inconsolata}

\usepackage{kotex}
\usepackage{graphicx}
\usepackage{booktabs}
\usepackage{csquotes}
\usepackage{dialogue}
\usepackage{lips}
\usepackage[para]{threeparttable}

\newcommand\frameworkname{WebTOD}

\title{Can Current Task-oriented Dialogue Models Automate\\ Real-world Scenarios in the Wild?\\}

\author{Sang-Woo Lee\thanks{\ \ Equal contribution.}\ $^{,1,2}$\ \ \  Sungdong Kim$^{*,2}$\ \ \  Donghyeon Ko$^{1}$\ \ \  Donghoon Ham$^{1}$\ \ \ \\
\textbf{Youngki Hong$^{1}$\ \ \ Shin Ah Oh$^{1}$\ \ \ Hyunhoon Jung$^{1}$\ \ \ Wangkyo Jung$^{1}$}\\ 
\textbf{ Kyunghyun Cho$^{3}$\ \ \ Donghyun Kwak$^{1}$\ \ \ Hyungsuk Noh$^{1}$\ \ \ Woomyoung Park$^{1}$}\\
\\
NAVER Cloud$^{1}$\ \ \  NAVER AI Lab$^{2}$\ \ \ NYU$^{3}$}

\begin{document}
\maketitle
\begin{abstract}
Task-oriented dialogue (TOD) systems are mainly based on the slot-filling-based TOD (SF-TOD) framework, in which dialogues are broken down into smaller, controllable units (i.e., slots) to fulfill a specific task. 
A series of approaches based on this framework achieved remarkable success on various TOD benchmarks.
However, we argue that the current TOD benchmarks are limited to surrogate real-world scenarios and that the current TOD models are still a long way to cover the scenarios.  
In this position paper, we first identify current status and limitations of SF-TOD systems.
After that, we explore the \frameworkname{} framework, the alternative direction for building a scalable TOD system when a web/mobile interface is available.
In \frameworkname{}, the dialogue system learns how to understand the web/mobile interface that the human agent interacts with, powered by a large-scale language model.
\end{abstract}
\section{Introduction}

Task-oriented dialogue (TOD) is one of the main research areas in dialogue modeling along with open-domain dialogue (ODD). TOD systems assist users in completing their goals, such as searching for information and making reservations. 

Most TOD systems are based on the slot-filling-based TOD (SF-TOD) framework~\cite{young-2013-pomdp}. 
SF-TOD assumes that a user has a specific goal, which could be represented by a set of slot-value pairs. The slot-value pairs of the goal correspond to a specific Application Programming Interface (API) to be executed for desired tasks. An SF-TOD system keeps asking the user for the values of missing or incomplete slots, to execute an API call properly. Slots, APIs, and a dialogue policy need to be predefined for each dedicated task by the developers of an SF-TOD system.

Most TOD studies that use large TOD benchmarks, such as MultiWOZ and SGD \cite{budzianowski2018multiwoz,rastogi2020towards}, fall under SF-TOD.
Since an SF-TOD system is dependent on the APIs which need specific and explicit information from the slots to perform a task, the benchmarks are based on the predefined slots, APIs, and a dialogue policy to support this constraint.
End-to-end TOD (E2E TOD) approaches have achieved state-of-the-art performances on these SF-TOD benchmarks~\cite{eric2017key, budzianowski2018multiwoz}. E2E TOD approaches have been proposed to generate all the intermediate information such as dialogue state, dialogue act, and final response, in an autoregressive manner \cite{ham2020end, hosseini2020simple, peng2021soloist, he2022galaxy}.

As SF-TOD requires developers to manually specify each task as slot-value pairs, corresponding APIs, and dialogue policy, it is challenging to implement an SF-TOD system for a complex real-world scenario.
It makes the use cases and applications of SF-TOD systems restricted to only a handful of simple cases that a few slots and simple API can easily represent.
In the simple cases, current SF-TOD systems are less competitive compared to other user interfaces such as the web/mobile as a solution to real-life tasks.
Due to these limitations, recent advances in SF-TOD \cite{he2022galaxy}, measured by a limited set of benchmarks, have not made into real-world TOD applications.

\begin{figure*}[t] 
\centering
\includegraphics[width=0.98\textwidth]{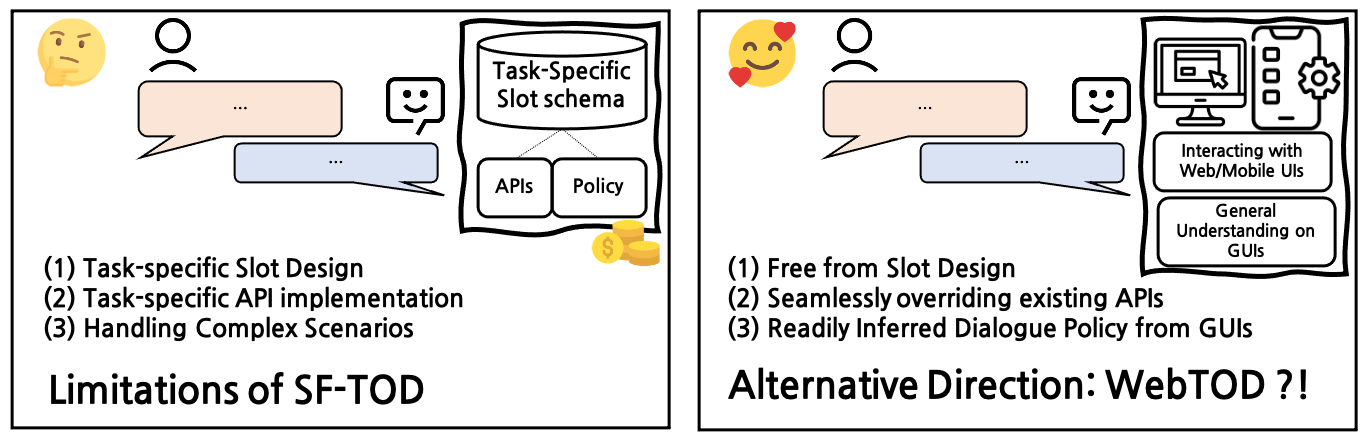}
\caption{Overview of the storyline for the limitations of SF-TOD and \frameworkname{}. Because of the limitations of SF-TOD, it needs much effort when encountering complex scenarios.
We conceptualize an alternative TOD framework, \frameworkname{} which can interact with the most popular user interfaces such as web/mobile to support tasks. Even though \frameworkname{} is not an ultimate solution, we believe it could resolve some of the limitations in SF-TOD in a way that requires less effort for implementing task-specific APIs and policies while understanding GUIs such as web/mobile.}
\label{fig:overview}
\end{figure*}

In this position paper, we discuss these issues in detail by describing the limitations of SF-TOD and then introducing an alternative idea which can be combined with already existing APIs easily.

The remaining parts of the paper are as follows. Section \ref{sec:slot-filling-based-tod} 
explains what SF-TOD is and what benchmarks are commonly used in SF-TOD.
Section \ref{sec:limitations-of-sf-tod} mainly identifies the limitations of SF-TOD systems. This section includes various interesting real-world scenarios. 
Section \ref{sec:behavioral-cloning-based-tod} conceptualizes \frameworkname{} framework, the alternative direction for scalable construction of TOD when a web/mobile interface is available.
The main storyline for the limitations of SF-TOD and \frameworkname{} is illustrated in Figure \ref{fig:overview}.

\medskip
Our contributions are as follows:
\begin{itemize}
    \item We discuss the limitation of SF-TOD, the mainstream trend in TOD research \cite{budzianowski2018multiwoz,rastogi2020towards}, describing the gap between TOD as a research field and TOD as a real-world solution.
    \item We conceptualize \frameworkname{} framework, which makes a general TOD model for web/mobile interfaces.
\end{itemize}
\section{Slot-filling-based TOD (SF-TOD)}
\label{sec:slot-filling-based-tod}

TOD aims to identify the user's veiled intention over multiple dialogue turns for accomplishing the user's goal. Different from ODD, it requires more precise control of the dialogue to achieve the goal. SF-TOD is the most representative and dominant approach to control agents for this purpose in the field. In this section, we describe SF-TOD and relevant benchmarks such as MultiWOZ and SGD~\cite{budzianowski2018multiwoz, rastogi2020towards}.

\subsection{SF-TOD}
\label{subsec:sf-tod}
SF-TOD assumes the user's goal can be characterized as a set of slot-value pairs, where the slots correspond to task-specific APIs. This assumption enables the precise control of the dialogue with the controllable units, i.e., slots. For example, the most straightforward way to operate the dialogue is to keep asking a user for values of slots that are not identified yet. 

To build an SF-TOD system, SF-TOD first needs to design APIs to perform the desired tasks. Each API contains function parameters and requires its function arguments to execute the API properly as a function. The slots and values correspond to the parameters and the arguments, respectively. Additionally, we can define what each slot is and the list of candidate values, in natural language. The slot specification is often called ``slot schema''~\cite{rastogi2020towards}. Finally, SF-TOD also needs to define a dialogue policy as well. The policy decides the proper next system action of the agent given a dialogue context.

\subsection{SF-TOD Benchmark}
\label{sec:sf-tod-benchmark}

The most representative benchmarks for the SF-TOD are MultiWOZ and SGD~\cite{budzianowski2018multiwoz, rastogi2020towards}. MultiWOZ is a multi-domain TOD dataset built on top of a single-domain TOD dataset, DSTC2~\cite{henderson-etal-2014-second}. It provides an evaluation protocol for both modular and end-to-end approaches. 

Many works have tested their models on multi-domain dialogue state tracking (DST), which is one of the core subtasks in SF-TOD \cite{wu-etal-2019-transferable, kim2020efficient}. DST evaluates whether a model correctly tracks the values of predefined slots that could be inferred from dialogue context. Joint goal accuracy (JGA) is used as a main evaluation metric for DST. It checks whether all predicted slot-value pairs are exactly matched with the ground-truth pairs at every dialogue turn. 
Similarly, SGD is proposed on the assumption of SF-TOD. Specifically, SGD evaluates the model's generalizability on unseen domains.

On the other hand, the MultiWOZ benchmark provides policy optimization and end-to-end modeling evaluations as well. The policy optimization evaluates the generated responses with a ground-truth dialogue state while the end-to-end modeling evaluates the responses generated from scratch, i.e., only conditioned on dialogue context. Inform and Success rates are used to measure dialogue-level task achievement~\cite{budzianowski2018multiwoz}. Specifically, the Inform rate checks whether the suggested DB entries by the system are within initial goal constraints and the Success rate checks whether the user's additional requests are answered well~\cite{nekvinda2021shades}.
Even though the benchmarks provide a quantitatively measurable proxy of well-performing TOD agents based on the predefined slot schema, we argue that these datasets are far from the real-world scenarios containing various complex cases in Section \ref{sec:limitations-of-sf-tod}.

\section{Limitations of SF-TOD}
\label{sec:limitations-of-sf-tod}

Section \ref{subsec:task-specific-slot-definition}, \ref{subsec:integration-of-existing-apis}, and \ref{subsec:complex-dialogue-policy-design} criticize the limitation of SF-TOD and its benchmark, describing problems that arise when SF-TOD framework is applied to complex real-world scenarios.

\subsection{Task-specific Slot Design}
\label{subsec:task-specific-slot-definition}

Most SF-TOD benchmarks assume that a user's goal can be fully described in a simple structure, for instance, as a set of slot-value pairs. This assumption allows a TOD system to easily control the dialogue via predefined slots. However, it does not consider cases beyond the assumption, for example, multiple values for a slot or the relationship among the slots. 
As a result, this assumption makes the system brittle even in simple cases. Consider the scenarios we provide below, in which this simple assumption does not fit.
We use U and S to denote the user and the system utterances, respectively.

\subsubsection{Single-Value Assumption}
\label{subsec:single-value-assumption}

Under a single-value assumption, a slot can take only one value \cite{ren-etal-2019-scalable}. Imagine a pizza ordering task, where ``menu-name'' is one of the predefined slots. Consider the following dialogue:

\begin{dialogue}
\speak{U} I would like to order \textbf{two pizzas}, a \textbf{Pepperoni} and a \textbf{Shrimp}.
\speak{S} Ok, what sizes do you want for each?
\end{dialogue}

This assumption makes it difficult to process the request since the ``menu-name'' can not take the two provided values, ``Pepperoni'' and ``Shrimp,'' at the same time.

\subsubsection{Single-Instance Assumption}
\label{subsec:single-instance-assumption}

Even if the single-value assumption is relaxed so that each slot may contain a list of values, rather than a singleton, it is not enough, as existing TOD systems often do not keep track of the relationship among multiple values across different slots. We refer to this limitation, or assumption of the lack of such relationship, as a ``single-instance assumption.'' In the following dialogue, we demonstrate how this limitation arises in the real world:

\begin{dialogue}
\speak{U} Regular size for both, please.
\speak{S} Ok, do you need anything else?
\speak{U} Oh, can I change the size of the \textbf{Pepperoni} pizza to a \textbf{large} size?
\end{dialogue}

In order to properly complete the order, the TOD system must be able to associate two different pizzas with two different sizes respectively. This can be understood as representing a goal not as a single set of slot-value pairs but as multiple sets of slot-value pairs~\cite{el-asri-etal-2017-frames}. 

\subsubsection{Simple-Value Assumption}
\label{subsec:realizing-the-slot-values}

Most SF-TOD benchmarks consider two types of slots. The first is a categorical slot, taking a value representing one of predefined categories. The other is a free-form slot in which the cases do not have predefined value candidates. In the latter case, the value of a slot is often treated as a string. However, both types can not leverage realistic constraints such as numerical or temporal operations.

\begin{dialogue}
\speak{U} ...
\speak{S} Booking was successful! You are reserved for  \textbf{16:30} on Saturday.
\speak{U} Oh I’m sorry but can you move it \textbf{back by two hours}?
\speak{S} No problem! Your reservation has been changed to \textbf{18:30}.

\end{dialogue}

For example, in the above example, the relative time expression, \textit{two hours later}, should be transformed into the operation `+' and the offset value $2 \times 60$. After then, the operation is performed on the base value (16:30), resulting in 18:30. 
The system obviously fails to identify the appropriate intention of users without a task-specific module for the operation.

It makes real-world TOD systems often require separate modules that operate slot values. However, implementing the additional modules considering such operations is costly.

\subsection{Task-specific API Implementation}
\label{subsec:integration-of-existing-apis}

The conventional SF-TOD approaches assume various formats of APIs and corresponding dialogue state representations. SF-TOD systems usually require dedicated APIs, but APIs dedicated to any specific TOD system rarely exist in advance and are often too costly to be implemented, especially in complex scenarios. Moreover, there are cases that are fairly simple for humans but are difficult for the system APIs to deal with.
We describe a case below.

\subsubsection{Customer Service Center Scenario}

Consider a scenario of a customer service center of an IT company that runs an e-mail service, in which a user with a blocked account requests an unblock. A human assistant responds to the request with a supporting GUI (API interaction block):

\begin{dialogue}
\speak{U} My email account seems to be blocked for being a spam account even though it isn't. Can you unblock it?
\speak{S} Hi, this is the customer service center. Let me check your account. Would you mind giving me a minute?
\par\small
\direct{\textbf{Start of API interaction}} \\
    > \textbf{\textit{Click Search User button}} \\
    > \textbf{\textit{Search user email account}} \\
    > \textbf{\textit{Click Account Summary button}} \\
    > \textbf\lips \\ 
    > \textbf{\textit{result: <html> (\dots) <value>1st Block</value> (\dots) </html>}} \\
    > \textbf{\textit{Click Unblock button}}\\
    > \textbf{\textit{Type ``Unblocking the first blocked user due to email request''}} \\
    > \textbf{\textit{Click Confirm button}} \\
\direct{\textbf{End of API interaction}}
\normalsize
\speak{S} Thank you for your patience. Since it’s your first time being blocked, I’ve unblocked your account without warning.
\end{dialogue}

The request in the above example depicts one of the popular requests in many real-life customer service centers and is actually not a difficult task for human agents. With the multiple sequential actions, however, this process is difficult for most conventional TOD systems. In the example above, the system would need to make the correct judgment (to unblock without warning) from the intermediate result, which is often not in a machine-friendly format (e.g., raw HTML text). Dedicated APIs can be implemented to solve this problem, but doing so for hundreds of tasks can be very expensive.

\subsection{Handling Complex Scenarios}
\label{subsec:complex-dialogue-policy-design}

The existing SF-TOD benchmarks deal with relatively simple scenarios, which are easy to handle in SF-TOD. However, it is difficult for SF-TOD to deal with complex scenarios, because complexity greatly affects the cost of designing a dialogue policy.

In this section, we describe realistic TOD scenarios requiring complex dialogue policy.
Section \ref{subsec:goal-clarification} shows that, in real-world situations, user queries can be much more complex than SF-TOD assumes. Section \ref{subsec:bad-user-experiences} generalizes the first example further, pointing out that SF-TOD is ignoring the UX perspective.

\subsubsection{Goal Clarification}
\label{subsec:goal-clarification}

Capturing a user goal becomes difficult when user goals are ambiguous.
Users sometimes begin conversations without even knowing themselves details necessary to complete their goals.

The following two scenarios are based on the same user goal of receiving a refund when a special discount coupon was used at the time of purchase. The user in Example 1 makes this request with a clear goal and enough details, while the user in Example 2 is not clear on whether a special discount coupon had been used for the purchase.

\begin{dialogue}
\speak{\textbf{Example 1}}
\speak{U} Will I get a full refund if I used the Special Monthly Sale coupon for the purchase?
\speak{S} Our policy is to issue an 80\% refund of the item’s price in that case.
\end{dialogue}

\begin{dialogue}
\speak{\textbf{Example 2}}
\speak{U} The refund I got is missing 2 dollars.
\speak{S} Did you purchase the item with a discount?
\speak{U} Hmm... I have no idea.
\speak{S} Let me check your purchase information.
\par\small
> Search the user’s purchase information
\normalsize
\speak{S} OK, I see. It looks like you used a coupon for the purchase. Can you remember what coupon it was?
\end{dialogue}

Normally, the system should be able to handle this particular user goal by classifying the situation as one of the known intents in the system. In the case of Example 2, however, the system cannot make such approach, as the user does not clearly state the goal and is not aware of key information. This can be overcome if slots that correspond to all such exceptional cases were provided, but it is intractable to include all the potential exception cases in the dialogue policy. 

\subsubsection{User Experience}
\label{subsec:bad-user-experiences}

An ideal dialogue system should provide a good user experience (UX) in various user scenarios with high accuracy, allowing the users to converse with the TOD system with comfort. Dialogue systems with good UX might be competitive as the system reduces the cognitive load of the users during the conversation. However, current literature on SF-TOD has focused on only joint goal accuracy and slot-filling/task success ratio as key metrics to optimize while ignoring the user experience aspect~\cite{budzianowski2018multiwoz, rastogi2020towards}. The metrics are easily aligned with predefined slots, and thus are good and tangible proxies to measure dialogue systems. But such metrics are likely to miss the crucial interactive nature of the dialogue system as described in Section \ref{subsec:goal-clarification}.
We believe a TOD system with good UX might handle complex scenarios that can not be represented easily with current slot designs.




\section{\frameworkname{}}
\label{sec:behavioral-cloning-based-tod}

\begin{figure}[t] 
\centering
\includegraphics[width=0.98\columnwidth]{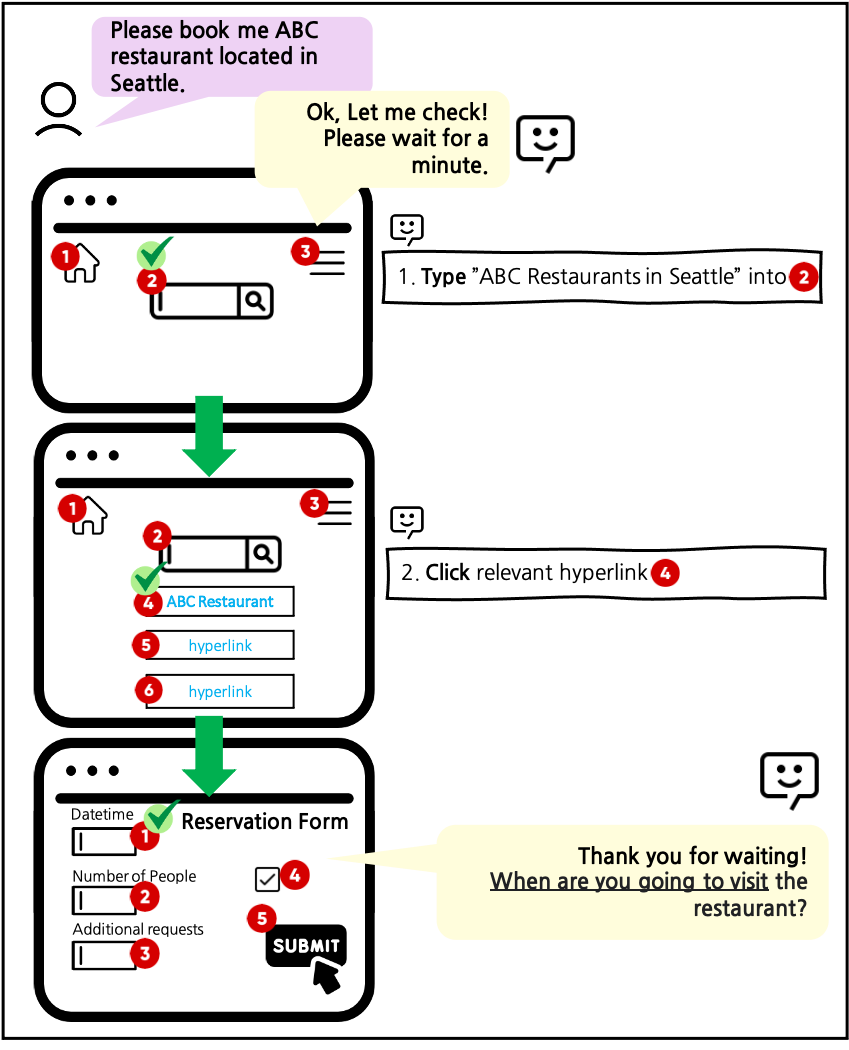}
\caption{Concept illustration of WebTOD. WebTOD can converse with users while naturally interacting with User Interfaces (UI) such as mobile apps and web browsers. The red dots indicate the actionable target elements in the current state of the UI. The action of WebTOD includes typing text into the text box and clicking buttons. Moreover, the actions are performed sequentially by necessity to proceed given task.}
\label{fig:webtod}
\end{figure}

This section introduces the concept of \frameworkname{} along with challenges and previous studies \cite{nakano2021webgpt,sun2022meta,2022act-1}. Imagine a TOD system that can interact with various types of graphic user interfaces (GUIs) such as web/mobile interfaces as illustrated in Figure~\ref{fig:webtod}.
By learning how to interact with elements such as buttons and text boxes on the web and mobile interfaces, the system can imitate a human agent by executing necessary actions on the web/mobile. \frameworkname{} is such a system that can execute actions to the GUIs and make a response based on the results of interactions with the interface, as a human agent uses GUIs to provide proper services. For example, \frameworkname{} could be integrated into the kiosk system in a restaurant to support older people who are unfamiliar with the kiosk system.

We start by explaining properties of \frameworkname{} in Section~\ref{subsec:bc-tod} and how WebTOD addresses the challenges in SF-TOD from the previous section in Section~\ref{subsec:general-tod--connection-to-challengs-in-sf-tod}. Then, we discuss the techniques that require further research along with important related works of \frameworkname{} in Sections~\ref{subsec:challenge-1-protocol-for-understanding-web-app} and \ref{subsec:challenge-2-pre-training-LLM}.

\subsection{Properties of \frameworkname{}}
\label{subsec:bc-tod}

We refer to \frameworkname{} as a framework for a system that has the ability to naturally interact with elements on the web/mobile, e.g., buttons and text boxes, to achieve the user's goal. We highlight two main properties of \frameworkname{} below.

\subsubsection{Interacting with Web/Mobile UIs}
\label{subsec:subapi-execution}

\frameworkname{} naturally navigates on any web/mobile interface by executing a sequence of actions. For example, it can click buttons, check boxes, or put texts in text boxes as well. 
This property makes the \frameworkname{} system scalable since it does not require developers to implement dedicated APIs for each task anymore. Moreover, \frameworkname{} could decide its action based on the trajectory of past actions and results from the interface, and conversational context with its opponents, i.e., users. 
An example of such a scenario with a web interface is depicted in Section \ref{subsec:integration-of-existing-apis}.

\subsubsection{General Understanding of GUIs} 
\label{subsec:general-understanding-on-gui}
The demo of ACT-1\footnote{
https://www.adept.ai/act
} \cite{2022act-1} presented by a startup called Adept is a motivating example for \frameworkname{}. In this demo, the system that surfs the web not only makes a reservation on the restaurant, but also can modify an Excel spreadsheet through a series of actions. 
Although it is unknown how far the actual development progress has been made, ACT-1's demo is a good example of the direction taken by \frameworkname{}, from the perspective of generalization on any form of web/mobile interfaces.

Humans can easily adapt and interact with any web/mobile interface even if they have never seen the interface before. However, it is not a trivial problem for machines to generalize their understanding over any web/mobile interface. We expect a large language model (LLM), such as GPT-3~\cite{brown2020language} could play a key role for \frameworkname{} by satisfying the generalization ability. For example, an LLM can take raw HTML text to understand the components of a web interface~\cite{aghajanyan2022cm3}. Another way is to understand the interface visually to get an HTML form~\cite{lee2022pix2struct}. We further discuss the challenge in Section~\ref{subsec:challenge-1-protocol-for-understanding-web-app}. 

We believe specialized pretraining might be required to tame the LLM for our purposes, e.g., a zero/few-shot understanding of elements in the unseen interface. For this, we can gather supervised training examples including action sequences over the environment as in \citet{nakano2021webgpt}. Moreover, we can leverage some indirectly related pretraining corpora to boost the performance of interface understanding. We discuss more pretraining corpora in Section \ref{subsec:challenge-2-pre-training-LLM}.

From this general GUI understanding, \frameworkname{} systems can interact with any interface and converse as a human agent as well, i.e., performing actions or generating proper responses based on past actions and results, and conversational context. Especially, in a case where a web/mobile interface implies a specific proper dialogue policy, there is no need for sophisticated dialogue policy design per each web/mobile interface. 

\subsection{Addressing the Challenges in SF-TOD}
\label{subsec:general-tod--connection-to-challengs-in-sf-tod}

In this section, we explain how \frameworkname{} addresses some of the challenges of SF-TOD.

\subsubsection{Free from Slot Design} 
Unlike the SF-TOD framework which requires to design task-specific slots, as described in Section \ref{subsec:task-specific-slot-definition}, \frameworkname{} systems can infer which slot is required by looking at the forms on the web/mobile interface. This makes task-specific slot design unnecessary.

\subsubsection{Seamlessly overriding existing APIs} 
In SF-TOD, it is necessary to design an API protocol for each task separately, as described in Section \ref{subsec:integration-of-existing-apis}. However, \frameworkname{} does not require this, as it can interact with the web/mobile interface implementing multiple actions based on a general understanding of GUIs. It means that \frameworkname{} is a scalable system as it seamlessly overrides already existing APIs on web/mobile interfaces in the real world without dedicated APIs implementations.

\subsubsection{Readily Inferring a Dialogue Policy from GUI} 
\label{subsec:more-policy-coverage-by-implicit-policy-information-in-gui}

As described in Section \ref{subsec:complex-dialogue-policy-design}, SF-TOD requires slot-based policy specification for each task.
However, \frameworkname{} can obtain most of the dialogue policy from GUIs by itself with fewer efforts, i.e., zero/few-shot dialogue policy learning.

Consider a case where we collect pretraining data from various web/mobile interfaces each of which corresponds to a complex scenario. Ideally, an LLM of \frameworkname{} can learn from these training data and derive a sophisticated and flexible dialogue policy from an arbitrary web/mobile interface. In this case, the \frameworkname{} system can provide a better UX experience to the user.

SF-TOD systems also can conduct a sophisticated dialogue policy with human effort for system implementation. However, as described in Section \ref{subsec:complex-dialogue-policy-design}, the implementation cost can be expensive. 



\subsection{Challenge 1: A Protocol for Understanding Web/Mobile Interface}
\label{subsec:challenge-1-protocol-for-understanding-web-app}

For current machine learning models, it is more challenging to understand general-domain web and mobile interfaces than to understand static documents that mainly consist of plain texts, such as Wikipedia articles. 
Here, we examine how we can translate web/mobile interfaces to a form more readily understandable to machine agents.

\subsubsection{Direction 1: Read GUI as Text}
\label{subsec:direction-1:read-gui-as-a-text}

To interact with the interface with only text, it is important to define effective text forms to understand and perform actions.

natbot is an opensource project about GPT-3 model controlling a web browser\footnote{Please check its demo at this \href{https://twitter.com/natfriedman/status/1575631194032549888}{url} and corresponding code at \href{https://github.com/nat/natbot}{github.com/nat/natbot}}. First, it parses the browser elements of a specific web page in text form. Link, image, button, and text box are examples of browser elements. Then, it inputs the list of parsed elements into InstructGPT~\cite{ouyang2022training} to make the model conduct actions on the web page to achieve a goal such as making a reservation at a restaurant. It allows limited actions for the control, e.g., scrolling, clicking, and typing and an element is one of the candidates for natbot to act on.
This project shows that a web parser with simple rules can read and parse a web page. An LLM can then read and act on the parsed list of items, as a TOD agent does.

\citet{sun2022meta} conduct a deeply related work of \frameworkname{}. They study how a machine can interact with a mobile app interface for six TOD domains, weather, calendar, search, taxi, restaurant, and hotel. Specifically, they detect actionable elements in the model app by extracting them from a layout file, which describes a structure of a specific UI of an Android app. The texts of extracted elements along with conversational history are fed into a model to predict the proper next action.

Current web/mobile parsing methods can be improved to cover a more diverse set of web and mobile pages.
For instance, the natbot parser can be improved by adding hierarchical information on its parsing form to make the LLM understand the hierarchy of elements in web/mobile pages.

\subsubsection{Direction 2: Using Visual Information}
Sometimes, the text does not contain enough information in terms of representing the state of GUIs.
We discuss two reasons here.
First, the web/mobile UIs consist of hierarchical elements and sometimes their structural information would be missed by parsing them as text sequences.
Second, some of the information on the web/mobile page is given as an actual image, which may include text. 
An example is an image-based advertisement that contains textual product information.

\citet{sun2022meta} also leverage visual features by obtaining elements-level image features from Faster R-CNN~\cite{ren2015faster}. The elements-level image features are fed into the transformer encoder along with other text features, e.g., dialogue history and textual information about elements. We believe recently proposed vision-language models such as CLIP~\cite{radford2021learning} make more performance improvements for this purpose.

On the other hand, OCR-free image-to-text models could also be explored for visual understanding of non-parsable, can not be understood by text format, elements in GUIs~\cite{kim2022ocr, lee2022pix2struct}.

\subsection{Challenge 2: Pretraining Corpora}
\label{subsec:challenge-2-pre-training-LLM}

Pretraining corpora related to the target task are insufficient to train \frameworkname{} in practice, because of the scarcity of the corpora. We think it could transfer its ability from a large unlabeled corpus indirectly related to \frameworkname{}, such as Web-crawled Text \cite{dodge2021documenting} and Github Code \cite{chen2021evaluating}.  
Also, conventional SF-TOD datasets also can be used for multi-task learning \cite{chen2022dialogzoo}. 
In addition to pretraining corpora, it will be helpful if there were browsing logs on the web/mobile interfaces, under the premise that there is no privacy issue \cite{shin2022pivotal}. 

Lastly, annotated datasets for supervised learning can be collected for \frameworkname{}. For all these types of datasets, the diversity in TOD scenarios will contribute to increased performance.

\section{Discussion}

\subsection{Where a Corresponding Web/Mobile Interface Does Not Exist}
Can \frameworkname{} paradigm be used effectively even when there is no corresponding web/mobile interface for the desired task?
In the case of various practical scenarios, the web/mobile interface might not exist, and a dialogue system must be implemented from scratch including the development of a corresponding interface.
However, we believe it is more considerable than the SF-TOD implementations when the dialogue scenario gets complex.
It can utilize the advantages from the perspective of data or UX, as described in Section \ref{subsec:more-policy-coverage-by-implicit-policy-information-in-gui}.

\subsection{Leveraging Human-to-Human Dialogues}
\label{subsec:leveraging-human-to-human-dialogues}
If there are target-domain conversation logs between a real human client and a human agent, can the conversation system be created based on these?
A good feature of \frameworkname{} is that \frameworkname{} can leverage these logs and app/web interfaces the agents use. Imagine there exist dialogue logs between human clients and human agents in target-domain real situations and action logs through an app/mobile interface created by human agents in the middle. These logs can be used for training the TOD model. 
The customer service center scenario introduced in Section \ref{subsec:integration-of-existing-apis} shows an example log.

Suppose that a customer service center collects enormous conversation and execution logs. It diminishes the need to collect training data and thus decreases the implementation cost of the TOD system.

\section{Related Works}

\subsection{Overcoming Limitations of SF-TOD}

To overcome the limitations discussed in Section~\ref{sec:limitations-of-sf-tod}, various approaches have been proposed. \citet{ren-etal-2019-scalable} discuss the problem of single-value assumption (Section \ref{subsec:single-value-assumption}) and propose a generative approach which tackles the assumption. 

To address the single-instance assumption (Section \ref{subsec:single-instance-assumption}), \citet{el-asri-etal-2017-frames} introduce a frame tracking task. Motivated by the exploring nature of the e-commerce domain, this task assumes a situation where multiple instances can achieve the user goal at the same time. The system then should keep track of multiple instances simultaneously and support natural switches and comparisons among the instances.

To handle numerical and temporal constraints (Section \ref{subsec:realizing-the-slot-values}), \citet{andreas-etal-2020-task} propose a dataflow graph-based representation to handle such complex user intents. Here, the graph is naturally extended as the conversation progresses, by resolving the constraints based on antecedents. Moreover, the graph representations are executable, and the next system response can be conditioned on the results of the execution, e.g., recovery from an exception raised by the execution.

\subsection{Browser-assisted Models}

\frameworkname{} is motivated by OpenAI WebGPT \cite{nakano2021webgpt}, which is an LLM for long-form question answering that searches the web and creates an answer based on multiple references. WebGPT uses OpenAI GPT-3 \cite{brown2020language} as a pretrained backbone and is further trained on the target task of long-form question answering. For finetuning, \citet{nakano2021webgpt} design a text-based web-browsing environment that the model can interact with. The environments accept actions such as searching with a query, clicking a link, finding text, quoting text, scrolling up and down, and producing an answer. Especially, by quoting text, the system can memorize important contents explicitly and use them by putting them in the prompt. The operations and environment's state are represented as a string which is provided to WebGPT as input. However, it is not scalable towards various GUIs since it requires specially designed environments for learning.

\subsection{Action on Web/Mobile}

To reduce the burden of developing dedicated APIs, universal executable semantic parsing and its corresponding programming languages have been studied~\cite{campagna2017almond, campagna2019genie}. These studies designed a programming language to make connections with several open APIs on the web. Recently,  executable semantic parsing has been successfully adapted to TOD benchmarks~\cite{lam2022thingtalk, campagna-etal-2022-shot}. It produces synthetic dialogues to train a contextual semantic parser.

 In another area of research, systems that directly perform a sequence of actions to various user interfaces have been studied, without relying on explicit executable semantics \cite{li2020mapping,gur2019learning,xu2021grounding}. In these studies, a user gives a text-based command to a system, and the machine executes a sequence of actions in an existing interface such as a mobile app and a web browser.

\section{Conclusion}
In this paper, we describe the limitations of SF-TOD and discuss new ideas to alleviate these issues.
We claim TOD systems based on these new ideas can be more competitive in complex dialogue scenarios.
We therefore urge TOD researchers to take more interest in dialogue systems which are more complex and real-world-oriented than the conventional SF-TOD systems. 
One of the most immediate future works is to design a new benchmark evaluating the challenges described in this paper, followed by developing an effective method for \frameworkname{}.
We hope to steer the current trend and discussion in TOD research toward more realistic TOD scenarios and challenges therein, in the future.

\section*{Limitations}

While we presented arguments and ideas with a certain stance in this paper, we did not suggest a specific benchmark or a detailed method. We hope that this paper would motivate and compel the community to begin tackling these real and important issues by developing relevant benchmarks and methods. 

We also did not cover all the existing domains and research directions in TOD. For example, negotiations \citep{lewis-etal-2017-deal}, one of the domains in TOD, is not discussed. \citet{zhang-etal-2022-toward-self} and \citet{geng-etal-2021-continual} deal with the problem of extending and improving the system agent in the online environment.
This is another important topic for the TOD systems in the real world where system maintenance and repairs are required, and can be applied in parallel with the methods mentioned in our paper.

\section*{Ethical Considerations}
TOD systems are especially in high demand in the industry and thus require extra ethical attention.

\subsection*{Privacy}
When the data is not guided by predefined schema and comes from real logs, the data becomes more uncontrollable and has more risk of including private information. For example, call logs of customer center, web browsing logs of users, and dialogue logs between a doctor and a patient are forms of dialogue data and have a risk of including private information. There is especially a high chance that the healthcare dialogue data includes sensitive information, and the misuse of private medical information can cause serious consequences \cite{bertino2005privacy, gostin2009beyond}. Thus, careful considerations are required when collecting such dialogue data. First, subjects of the data should be provided with a proper alert so that they are aware of the fact that the data is being collected. Second, if required, experimental design must be reviewed by a proper ethics reviewer, such as Institutional Review Board \cite{valizadeh2022ai}. Lastly, privacy-preserving data collection methods such as de-identification need to be considered \cite{shim2021building}.

\subsection*{Bias}
It is also well-known that large, uncontrolled data can suffer from undesirable biases involving gender, race, or religion \cite{bolukbasi2016man, bordia-bowman-2019-identifying, bartl-etal-2020-unmasking, shah-etal-2020-predictive}, as exemplified by the case of Microsoft Tay Chatbot \cite{neff2016talking}. The dialogue system that trains the data with biases can generate responses that contain the biases. There are many methods that attempt to remove the biases from the data \cite{bolukbasi2016man, tripathi2019detecting}. 

\subsection*{Misdirection}
TOD systems mostly target to be served as consumer products. Therefore, if the system gives incorrect information and misdirects users, it can harm its users (e.g., financial damage or psychological distress). If the system deals with healthcare domain, which is one of the common target domains of TOD systems, gives incorrect predictions, it can even result in serious medical accidents \cite{valizadeh2022ai}. There should be some countermeasures to complement the imperfect performance of the system. It can be applying strict threshold to the system's response so that the response is generated only when the system is confident about the response, or providing an alert that warns about imperfectness of the system.

\section*{Acknowledgements}
The authors thank all the members of NAVER Cloud and AI Lab for their devoted support. In particular, they thank Nako Sung, Jung-Woo Ha, Sohee Yang, Hwaran Lee, and Joonsuk Park for intense discussion and proofreading.

\bibliography{custom}
\bibliographystyle{acl_natbib}

\clearpage
\appendix

\end{document}